\documentclass{article}
\usepackage{arxiv}
\usepackage[utf8]{inputenc}
\usepackage[T1]{fontenc}
\usepackage{graphicx}
\usepackage{float}
\usepackage[ruled,vlined]{algorithm2e}
\usepackage[detect-all]{siunitx}
\usepackage{amsmath, multicol, amssymb}
\usepackage{xcolor}

\definecolor{customGreen}{RGB}{0,153,0}

\title{Reinforcement Learning for Assignment problem}

\author{
	Filipp Skomorokhov \\
	Moscow Institute of Physics and Technology,\\ Skolkovo Institute of Science and Technology\\
	\And
	George Ovchinnikov \\
	Skolkovo Institute of Science and Technology\\
}

\begin{document}
\maketitle

\begin{abstract}	
This paper is dedicated to the application of reinforcement learning combined with neural networks to the general formulation of user scheduling problem. Our simulator resembles real world problems by means of stochastic changes in environment. We applied $Q$-learning based method to the number of dynamic simulations and outperformed analytical greedy-based solution in terms of total reward, the aim of which is to get the lowest possible penalty throughout simulation.
\end{abstract}

\section{Introduction}
On Demand services, such as a ride sharing \cite{ridesharing}, coordination of multiply robots \cite{coveragecontrol}, user serving in MIMO networks \cite{PF_MIMO} etc utilize  management strategies in order to improve customer quality of service (QoS) requirements.
The problem of shared resource utilization is very common in wireless networks \cite{overviewMIMO}  and becoming more important with more devices connected because of development of IoT and 5G.

Usually such systems have multiply concurrent users awaiting serving and fewer number of workers resources available, along with switching costs from serving user to user (like trip for taxi driver from drop off of one user to pick up point of the next one). Real world systems are dynamic in nature with cause and effect information not being given and system behavior and QoS only being observed.

Previous works developed different algorithmic or classical scheduling methods, where QoS is maintained via algorithm using some sort of priority index, like Proportional Fair \cite{simpleQoS}, \cite{PF_MIMO} or MLWDF \cite{MLWDF}.  
This work focuses on reinforced learning applied to general formulation of user scheduling problem. 
A $Q$-learning based method is presented for maximizing customer QoS and compared to analytical strategies. A $Q$-learning approach is shown to improve QoS up to TODO$\%$ compared to baseline scenarios.

\section{Problem statement}
We propose a simple and general model of resource planing (or user scheduling). It can be described as following:

We consider a number of users for each there is a fixed beforehead (but possibly different for each user)  amount of data to be transmitted. 

Transmission happens in the time steps and only a certain amount of data, 
depended on user 
can be transmitted during the time step. 
This amount is defined by the weight vector $w$ of size $N$ where $N$ is the number of users. 

After each user is done, we decide which of the users will be served next.
Simulation is considered finished when there is no data to be send for all users. 

There are switching costs, so when we choose a user $i$, we give a penalty for this user interfering with all users, previously served during $m$ timesteps, in our simulations $m=2$. The set of these users is denoted by $M$. The size of individual penalty is given according to the penalty matrix $P$ ($N$ x $N$, $N$ is the number of users) the following way: 
\begin{equation}
p = \sum_{k\in M} P_{ik}. \label{penalty}
\end{equation}

The aim of the agent is to minimize total penalty throughout the simulation.

\section{Experiments description}
Our simulator consists of 1 transmitter and 10 Users. 
We want to minimize penalty \eqref{penalty}, but for convenience we want to maximize our cumulative reward, 
by setting penalty as negative numbers, so we are searching for optimal scheduling strategy $\pi^*$ which gives the ordering of the users with maximum reward:
$$\pi^* = \arg \max_{\pi} \sum_{i} P_{\pi_i}.$$

Our state consists of buffer remainders - amount of work left to be done $s_1 \in \mathbb{R}^N$ and vectorized penalty matrix $P \in \mathbb{R}^{N^2}$. So totally we use state  $S \in \mathbb{R}^{N+N^2}$. Taking into consideration that our simulations use 10 Users, we have $S \in \mathbb{R}^{110}$.

And, as it was mentioned earlier, our action $\mathcal{A} \in \mathbb{R}$ is descrete in range \numrange{1}{10}.

\section{Baseline}
For our baseline we chose the algorithm, which acts in a greedy manner. At each timestep $t$ a user $u_t$ with the highest possible reward $r_t$ at $t$ is chosen.

\begin{algorithm}[]
$t$ = 1 \;
Initialize $s_t$ from simulator \;
\While{$s_t$ is not a terminal state}{
	$u_t$ $\leftarrow \arg\max_{k}{ P_{k, u_{t-1}} +  P_{k, u_{t-2}}}$ \;
	Make a simulation step with $u_t$, receiving $s_{t+1}$ \;
	$s_t \leftarrow s_{t+1}$ \;
	$t \leftarrow t+1$
}
\caption{Baseline: Greedy policy} \label{alg:baseline}
\end{algorithm}

The downside of the baseline is that it is not able to take actions based on some multi-steps strategy. And this is difficult since we have dynamics in our system. As weight vector $w$ and penalty matrix $P$ change over time, their instant values are not enough to develop a long-term strategy. But, as we will show later, such a strategy can be found by our agent and one of the reasons for this is that dynamics we mentioned earlier is, of course, not completely random.

\section{$Q$-learning and setup}
The main idea behind $Q$-learning from \cite{Sutton1998} is that if we had a function $Q(s, a)$, that could tell us what our return would be, if we were to take an action in a given state, then we could easily construct a policy that maximizes our rewards:
$$\pi^{*}(s)=\operatorname{argmax}_{a} Q^{*}(s, a)$$
However, we don’t know everything about the environment, so we don’t have access to $Q^{*}$. Moreover, classical approach would tell us to build a table of state-action values with the size of all possible combinations of states and actions. Frequently, in real world tasks this number can be extremely huge for us to operate with this table. That's why we switch to Deep Reinforcement Learning approach, which implies using a neural network as a great function approximator. So we can create one and train it to resemble $Q^{*}$.

\pagebreak

For our training update formula, we’ll use a fact that $Q$ function for some policy obeys the Bellman equation:
$$Q^{\pi}(s, a)=R+\gamma Q^{\pi}\left(s^{\prime}, \pi\left(s^{\prime}\right)\right)$$
Here $R$ is a reward, received from taking action $a$ from state $s$.
The difference between the two sides of the equality is known as the temporal difference error (TD-error), $\delta$:
$$\delta=Q(s, a)-\left(R+\gamma \max _{a} Q\left(s^{\prime}, a\right)\right)$$

To add more stability to the training process, TD-error uses not single neural network $Q$, but two networks: policy and target networks. This stabilization technique was introduced in \cite{mnih2015humanlevel}. \\
Policy network is used to sample actions from it, it's weights are updated at each optimization iteration. Target network is different in sense that it's weights are updated every $upd$ steps. In TD-error it's used under \texttt{max} operator.

So now TD-error is:
$$\delta=Q_{policy}(s, a)-\left(R+\gamma \max _{a} Q_{target}\left(s^{\prime}, a\right)\right)$$

$Q$-learning is an off-policy algorithm, this means that we can use experience replay first introduced in \cite{mnih2015humanlevel} to increase stability of training process and sample efficiency by reusing sample tuples $\{s, a, r, s^{\prime}\}$. We need this to speed up experiment, because simulation process is quite slow. The procedure of sample collecting lies in running multiple sample gathering simulations. When these simulations finish, we integrate all samples to one huge base of samples. Later, during training process, we don't use simulation, but our samples only. No simulation training idea was achieved during one of experiments, so not all of them use this idea.

\begin{algorithm}[H]	
	
Algorithm parameters:
\begin{itemize}
	\item Discounting factor $\gamma \in [0, 1]$
    \item Target update $upd$
\end{itemize}

Initialize $Q_{policy}$ weights \;
$Q_{target} \leftarrow Q_{policy}$ \;
Initialize $s$ from simulator \;
$t$ = 1 \;

\ForEach{$\{s, a, r, s'\}$ from samples}{
    $\delta=Q_{policy}(s, a)-\left(r+\gamma \max _{a^{\prime}} Q_{target}\left(s^{\prime}, a^{\prime}\right)\right)$\;
    $Update \; Q_{target} \; from \; \delta$ \;
    \uIf{if $t \bmod upd = 0$}{
        $Q_{target} \leftarrow Q_{policy}$ \;
    }
    $s \leftarrow s'$\;
    $t \leftarrow t + 1$ \;
}
\caption{$Q$-learning}
\end{algorithm}

\section{Neural network architecture}
\begin{figure}[H]
	\centering
	\includegraphics[width=0.7\textwidth]{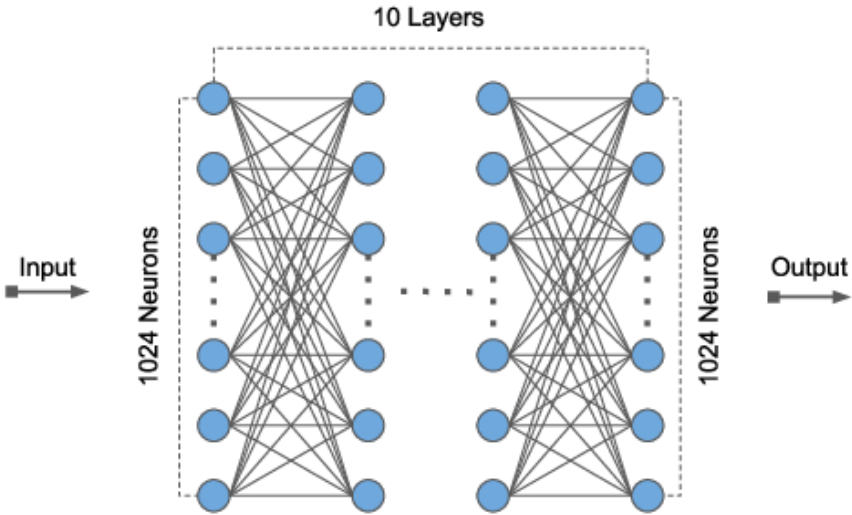}
	\caption{Neural network architecture}
	\label{fig:nn_architecture}
\end{figure}

Of course, the main thing we wanted from neural network was to complement $Q$-learning algorithm itself in ability of capturing strategies, which are effective in terms of total reward. The final architecture of the network can be described as follows (also shown on figure \ref{fig:nn_architecture}): 10 Fully-Connected layers with 1024 neurouns at each and ReLU as activation function. Our state representation consists of vectorized matrices of real numbers, this means that the main constructive block in our neural network is Fully-Connected layer.

\section{Results}
\begin{figure}[H]
	\centering
	\includegraphics[width=0.5\textwidth]
	{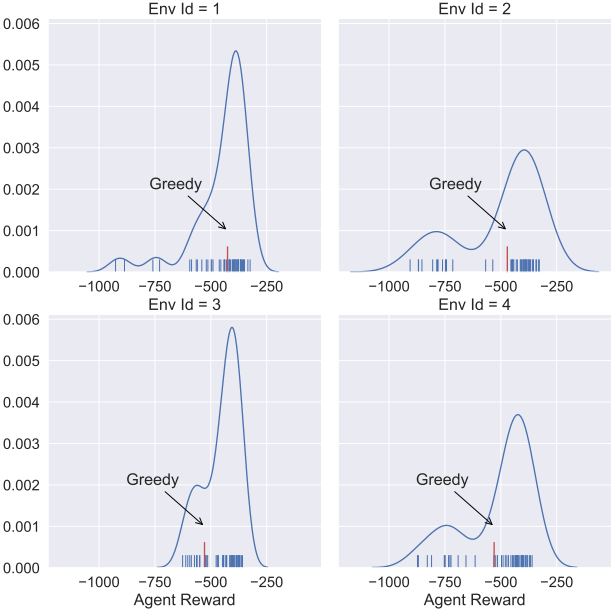}
	\caption{Distribution of agents}
	\label{fig:agent_distr} 
\end{figure}

Our results are presented at the graphs below. The first one, figure \ref{fig:agent_distr}, shows distribution of different agents (trained with different random seeds) per environment built with kernel density estimation (KDE). There is also greedy algorithm \ref{alg:baseline} shown for each environment. For convenience we present 4 environments here. Full graph can be found in appendix (section \ref{appendix_A}). X axis shows total simulation reward. As we mentioned earlier, it is a negative number for the purpose of convenience. On y axis we show probability density. As we see from this graph, the majority of our agents outperformed baseline (greedy algorithm) in each environment.

As it was said, the dynamics of the environment parameters $w$ and $P$ is not random. They change in the manner of Brownian movement. And environments are different in terms of the parameters of this distribution. The majority of agents were able to capture or at least successfully use this dependency to maximize their total reward. 

Now we present an aggregated table:

\begin{table}[H]
\centering
\begin{tabular}{ |p{1.8cm}||p{1.8cm}|p{1.8cm}|p{1.8cm}|  }
	\hline
	\multicolumn{4}{|c|}{Greedy and Best Agent results} \\
	\hline
	Environment ID& Greedy result &Best Agent result&Advantage of Agent\\
	\hline
	1 & -425 & -324 &  \leavevmode\color{customGreen} +31\% \\
	2 & -471 & -328 &  \leavevmode\color{customGreen} +43\% \\
	3 & -528 & -358 &  \leavevmode\color{customGreen} +47\% \\
	4 & -530 & -360 &  \leavevmode\color{customGreen} +47\% \\
	5 & -437 & -335 &  \leavevmode\color{customGreen} +30\%\\
	6 & -480 & -330 & \leavevmode\color{customGreen} +45\%\\
	7 & -515 & -389 & \leavevmode\color{customGreen} +32\%\\
	8 & -456 & -356 & \leavevmode\color{customGreen} +28\%\\
	9 & -379 & -342 & \leavevmode\color{customGreen} +10\%\\
	10 & -405 & -352 & \leavevmode\color{customGreen} +15\%\\
	11 & -435 & -326 & \leavevmode\color{customGreen} +33\%\\
	12 & -525 & -355 & \leavevmode\color{customGreen} +47\%\\
	13 & -367 & -313 & \leavevmode\color{customGreen} +17\%\\
	14 & -458 & -356 & \leavevmode\color{customGreen} +28\%\\
	15 & -509 & -346 & \leavevmode\color{customGreen} +47\%\\
	16 & -490 & -320 & \leavevmode\color{customGreen} +53\%\\
	\hline
\end{tabular}
\caption{Best agents, baseline and advantage results}\label{tab:a}
\end{table}

Last column shows the difference between best Agent and greedy-based baseline for certain environment in percents. Positive number here indicates that the agent is better than baseline.

So the whole table \ref{tab:a} shows that for all environments there are agents which outperformed baseline. Advantage varies from 10\% (Env ID $= 9$) to 53\% (Env ID $= 16$).

\begin{figure}[H]
	\centering
	\includegraphics[width=0.7\textwidth, scale=1.0]{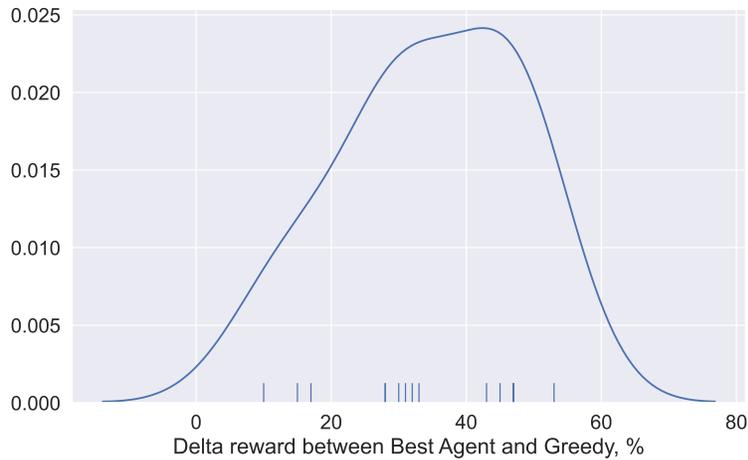}
	\caption{Distribution of advantages of best agents per environment}
	\label{fig:total_distr}
\end{figure}

In order to make this result more vivid, the second graph (figure \ref{fig:total_distr}) shows it as a distribution of advantage. Each point here is delta reward (in percents) between greedy algorithm and best agent for each environment, just like in table \ref{tab:a}.

The main point from this graph is that some average advantage can be nearly 30 to 50\%, which allows us to say that reinforcement learning can successfully be applied to the user scheduling problem \eqref{penalty}.

\section{Conclusion}
The main result of our paper is that successful appliance of Reinforcement Learning methods ($Q$-learning, in our case) to the general formulation of user scheduling problem \eqref{penalty} is possible. This means that for various environments we found an agent, which outperformed the baseline (algorithm \ref{alg:baseline}) with a certain margin, described in table \ref{tab:a}. And this result along with the presence of stochasticity in main environment parameters such as penalty matrix $P$ and weight vector $w$ is a good sign that our approach has ability to capture or use dependencies for total reward maximization and our results can be developed further into different real world applications.

\clearpage
\section{Appendix A: Distribution of agents for all environments}
\label{appendix_A}
\begin{figure}[H]
\centering
\includegraphics[width=0.55\textwidth]{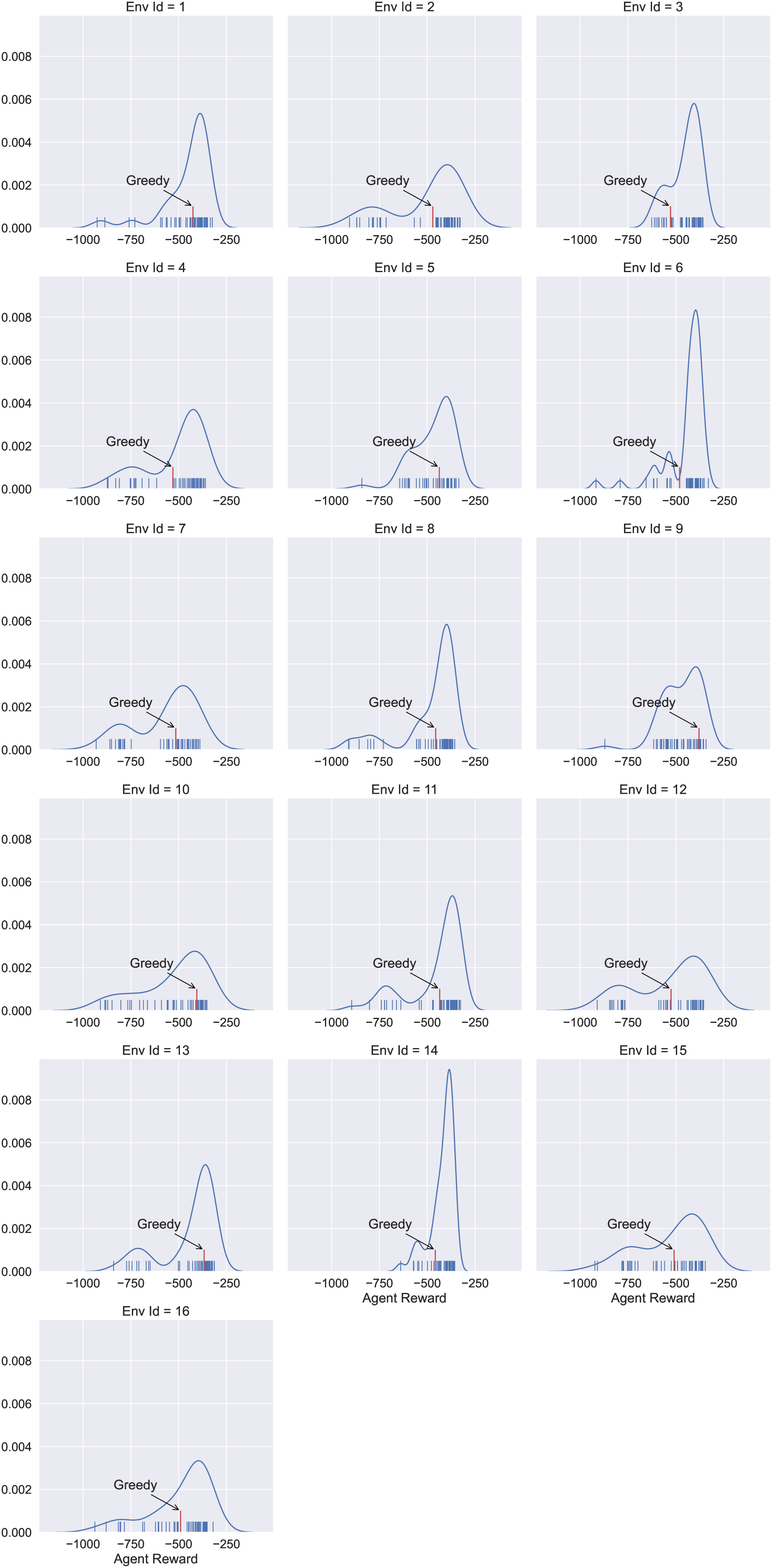}
\caption{Distribution of agents for all environments}
\label{fig:distr_full}
\end{figure}

\bibliographystyle{unsrt}

\begin{thebibliography}{1}

\bibitem{ridesharing}
J.~{Miller} and J.~P. {How}.
\newblock Predictive positioning and quality of service ridesharing for campus
  mobility on demand systems.
\newblock In {\em 2017 IEEE International Conference on Robotics and Automation
  (ICRA)}, pages 1402--1408, 2017.

\bibitem{coveragecontrol}
A.~{Sadeghi} and S.~L. {Smith}.
\newblock Coverage control for multiple event types with heterogeneous robots.
\newblock In {\em 2019 International Conference on Robotics and Automation
  (ICRA)}, pages 3377--3383, 2019.

\bibitem{PF_MIMO}
Lingjia Liu, Young-Han Nam, and Jianzhong Zhang.
\newblock Proportional fair scheduling for multi-cell multi-user mimo systems.
\newblock In {\em 2010 44th Annual Conference on Information Sciences and
  Systems (CISS)}, pages 1--6. IEEE, 2010.

\bibitem{overviewMIMO}
Hossam Fattah and Cyril Leung.
\newblock An overview of scheduling algorithms in wireless multimedia networks.
\newblock {\em IEEE wireless communications}, 9(5):76--83, 2002.

\bibitem{simpleQoS}
{Wai-keung Fung}, {Ning Xi}, {Wang-tai Lo}, and {Yun-hui Liu}.
\newblock Improving efficiency of internet based teleoperation using network
  qos.
\newblock In {\em Proceedings 2002 IEEE International Conference on Robotics
  and Automation (Cat. No.02CH37292)}, volume~3, pages 2707--2712 vol.3, 2002.

\bibitem{MLWDF}
Matthew Andrews, Krishnan Kumaran, Kavita Ramanan, Alexander Stolyar, Phil
  Whiting, and Rajiv Vijayakumar.
\newblock Providing quality of service over a shared wireless link.
\newblock {\em IEEE Communications magazine}, 39(2):150--154, 2001.

\bibitem{Sutton1998}
Richard~S. Sutton and Andrew~G. Barto.
\newblock {\em Reinforcement Learning: An Introduction}.
\newblock The MIT Press, second edition, 2018.

\bibitem{mnih2015humanlevel}
Volodymyr Mnih, Koray Kavukcuoglu, David Silver, Andrei~A. Rusu, Joel Veness,
  Marc~G. Bellemare, Alex Graves, Martin Riedmiller, Andreas~K. Fidjeland,
  Georg Ostrovski, Stig Petersen, Charles Beattie, Amir Sadik, Ioannis
  Antonoglou, Helen King, Dharshan Kumaran, Daan Wierstra, Shane Legg, and
  Demis Hassabis.
\newblock Human-level control through deep reinforcement learning.
\newblock {\em Nature}, 518(7540):529--533, February 2015.

\end{thebibliography}

\end{document}